\documentclass[letterpaper]{article} 
\usepackage{aaai25}  
\usepackage{times}  
\usepackage{helvet}  
\usepackage{courier}  
\usepackage[hyphens]{url}  
\usepackage{graphicx} 
\usepackage{natbib}  
\usepackage{caption} 
\def\ourmodel{PmNet}
\def\ourdata{PmLR50}
\usepackage{amssymb}
\usepackage{amsmath}
\usepackage{multirow}
\usepackage{algorithm}
\usepackage{algorithmic}
\usepackage{xspace}
\usepackage{amsmath,bm}
\usepackage{color}
\usepackage{colortbl}
\usepackage{subfigure}
\usepackage{pifont}
\usepackage{xcolor}
\usepackage{newfloat}
\usepackage{listings}
\usepackage{bbding}

\newcommand{\cmark}{\ding{51}}%
\newcommand{\xmark}{\ding{55}}%
\newcommand{\tabref}[1]{Table \ref{#1}}

\newcommand{\figref}[1]{Fig. \ref{#1}}

\def\eg{\emph{e.g.}}
\def\ie{\emph{i.e.}}

\usepackage{newfloat}
\usepackage{listings}
\DeclareCaptionStyle{ruled}{labelfont=normalfont,labelsep=colon,strut=off} 
\floatstyle{ruled}
\newfloat{listing}{tb}{lst}{}
\floatname{listing}{Listing}
\title{Surgical Workflow Recognition and Blocking Effectiveness Detection \\ in Laparoscopic Liver Resections with Pringle Maneuver}
\author{
    Diandian Guo\textsuperscript{\rm 1}\equalcontrib,
    Weixin Si\textsuperscript{\rm 2}\equalcontrib,
    Zhixi Li\textsuperscript{\rm 3},
    Jialun Pei\textsuperscript{\rm 1}\thanks{Corresponding author.},
    Pheng-Ann Heng\textsuperscript{\rm 1}
    }
\affiliations{
    \textsuperscript{\rm 1}Department of Computer Science and Engineering,
    The Chinese University of Hong Kong\\
    \textsuperscript{\rm 2}Shenzhen Institute of Advanced Technology, 
    Chinese Academy of Sciences\\
    \textsuperscript{\rm 3}Nanfang Hospital, Southern Medical University\\
    {jialunpei@cuhk.edu.hk}}

\begin{document}

\maketitle

\begin{abstract}
Pringle maneuver (PM) in laparoscopic liver resections aims to reduce blood loss and provide a clear surgical view by intermittently blocking blood inflow of the liver, whereas prolonged PM may cause ischemic injury.  
To comprehensively monitor this surgical procedure and provide timely warnings of ineffective and prolonged blocking, we suggest two complementary AI-assisted surgical monitoring tasks: workflow recognition and blocking effectiveness detection in liver resections. The former presents challenges in real-time capturing of short-term PM, while the latter involves the intraoperative discrimination of long-term liver ischemia states. 
To address these challenges, we meticulously collect a novel dataset, called \textbf{\ourdata}, consisting of 25,037 video frames covering various surgical phases from 50 laparoscopic liver resection procedures. Additionally, we develop an online baseline for~\ourdata, termed \textbf{\ourmodel}. This model embraces Masked Temporal Encoding (MTE) and Compressed Sequence Modeling (CSM) for efficient short and long-term temporal information modeling, and embeds Contrastive Prototype Separation (CPS) to enhance action discrimination between similar intraoperative operations. Experimental results demonstrate that~\ourmodel~outperforms existing state-of-the-art surgical workflow recognition methods on the~\ourdata~benchmark. Our research offers potential clinical applications for the laparoscopic liver surgery community.
\end{abstract}
\begin{links}
    \link{Code}{https://github.com/RascalGdd/PmNet}
\end{links}
\section{Introduction}
With the advancement of minimally invasive surgery, laparoscopic procedures have gained widespread acceptance among surgeons with advantages such as smaller incisions, reduced pain, and faster recovery~\cite{Sidaway2024,Lu2024}. 
However, the limited surgical field and complex surrounding environment increase the risks associated with intraoperative actions~\cite{Okamura2019}. 
Particularly in liver resections, it is critical to reduce blood loss and maintain a clear operative view during surgery due to the complex vascular anatomy of the liver. 
Pringle maneuver (PM), a regular and gold standard technique in laparoscopic liver resections, involves using the Foley catheter or Nylon urinary catheter to clamp the hepatoduodenal ligament, thereby intermittently blocking blood inflow to the liver~\cite{Khajeh2021,man1997prospective}. 
However, prolonged blocking may cause liver ischemic injury leading to liver dysfunction, while ineffective blocking increases the risk of intraoperative bleeding and mishandling due to unclear surgical view. 
In this regard, comprehensive monitoring and early warning for this high-risk surgical procedure are essential to enhance the safety of liver resections and alleviate the decision-making pressure on surgeons.

\begin{figure}[t]
	\centering
	\includegraphics[scale=.172]{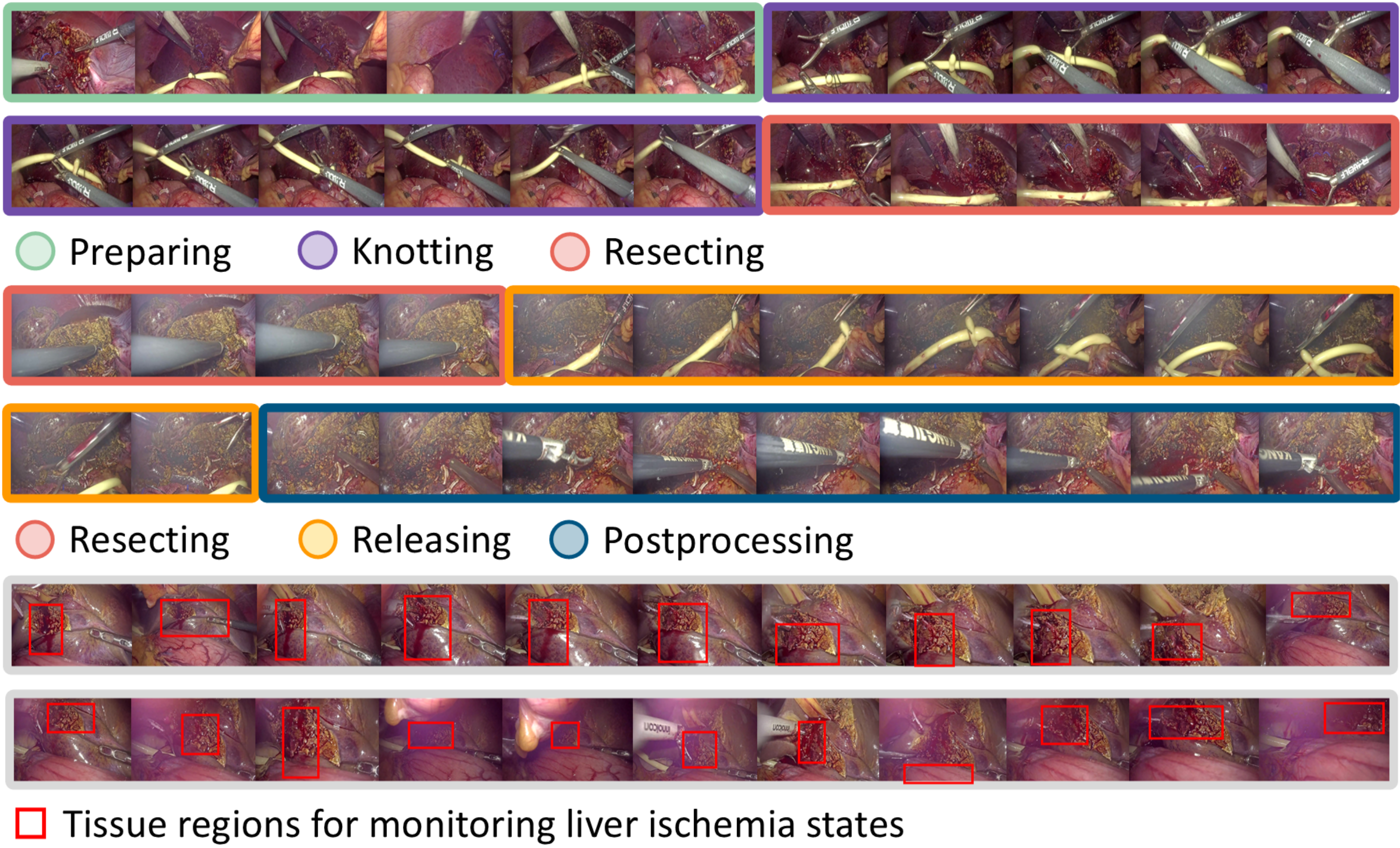}
	\caption{Overview of two typical features of surgical workflows in laparoscopic liver resections with Pringle maneuver in~\ourdata. (1) The Knotting and Releasing operations are relatively rapid and similar in the entire surgical procedure. (2) Blocking effectiveness detection involves long-term monitoring of liver ischemic states and bleeding of blocking-irrelevant frames, which have little impact on short-term operation integration. Zoom in for details.}
	\label{start}
\end{figure}

AI-assisted surgical workflow recognition has proven effective in intraoperative decision-making and workflow optimization~\cite{Cao2023,last,lessismore,weaklysupervised}. However, while numerous excellent methods exist for surgical workflow recognition~\cite{deepphase,svrcnet,skit,lovit}, research dedicated to monitoring high-risk liver resection procedures remains limited, especially on process monitoring and early warning of PM. 
In contrast to other surgical workflow recognition tasks, liver resections require real-time intraoperative monitoring of long-term ischemic states and liver surface color changes after blocking, as well as assessing the effectiveness of inflow occlusion. 
Further, capturing and differentiating two rapid and easily confused operations (\eg, Knotting and Releasing) of PM in long-term surgical videos poses a unique challenge for this task.

Although various surgical workflow datasets~\cite{autolaparo,micai16,heichole,cataract101,thadataset} already exist in the field of surgical video analysis, there are very few publicly available datasets for liver resection surgery.
To advance research on workflow recognition and blocking effectiveness detection in liver resections , we construct a brand-new surgical workflow recognition dataset named~\ourdata, which focuses on PM procudeures in liver resections. 
Our dataset comprises 50 surgical cases with 25,037 video frames. All samples are collected and annotated for phase classification and effectiveness binary labels by six hepatobiliary surgeons. 
As illustrated in~\figref{start}, the procedure of PM consists of five phases: \emph{Preparing, Knotting, Resecting, Releasing}, and \emph{Postprocessing}. 
Besides, we provide auxiliary bounding boxes for monitoring liver ischemic states to assist in detecting blocking effectiveness. 
Building on the~\ourdata~benchmark, we propose two complementary tasks: PM workflow recognition and blocking effectiveness detection. The former presents a challenge for the capture of both long- and short-term surgical operations, while the latter involves modeling long-term ischemic states and subtle color changes of the liver during surgery.

To address the above challenges, we propose a unified online baseline called~\ourmodel~for both PM workflow recognition and blocking effectiveness detection. 
Our framework includes a Masked Temporal Encoding (MTE) that emphasizes surgical action details by adaptively masking and swapping message tokens for efficient short-term temporal information modeling. Moreover, we introduce a Compressed Sequence Modeling (CSM) operation, which leverages the state space model (SSM) to create long-range dependencies on temporally pooled long-term features and information of tissue ischemia regions for compressed feature modeling, and then interacts short-term memory with long-term memory to facilitate contextual retrieval. 
Additionally, our model embeds a Contrastive Prototype Separation (CPS) strategy to expand the feature space distance across target prototypes, improving the discrimination between similar intraoperative actions. We comprehensively evaluate multiple mainstream surgical workflow recognition methods to construct the~\ourdata~benchmark. 
Experimental results demonstrate that~\ourmodel~outperforms previous state-of-the-art methods on the~\ourdata~test set, proving potential clinical significance in surgical intervention and assistance. Our main contributions are summarized as:

\begin{itemize}
\item We first present the surgical workflow recognition as well as blocking effectiveness detection task geared towards liver resections with PM. Accordingly, we collect a novel dataset called~\ourdata~and establish a comprehensive benchmark to facilitate relevant community study.
\item We propose an online baseline, termed~\ourmodel, for PM workflow recognition together with blocking effectiveness detection. This method can efficiently perform long-term and short-term memory temporal modeling for real-time workflow recognition and judgment of blocking effectiveness, achieving superior performance on the~\ourdata~benchmark.
\item Masked Temporal Encoding (MTE) is introduced to adaptively filter out blocking-irrelevant features for effective aggregation of short-term temporal dynamics. Besides, a Mamba-based Compressed Sequence Modeling (CSM) is designed to model compressed features on temporally pooled long-term information, facilitating the interaction between short- and long-term memories.
\item We propose Contrastive Prototype Separation (CPS), a contrastive learning strategy that extends the feature space between confusing surgical operations for the accurate understanding of intraoperative actions.
\end{itemize}
\begin{figure*}[t]
	\centering
	\includegraphics[scale=.266]{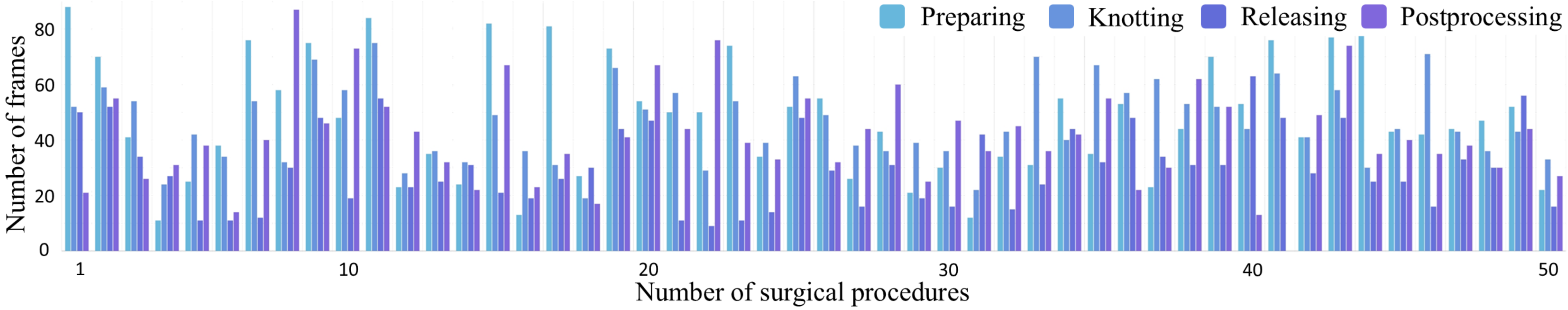}
	\caption{\textbf{Histogram.} PM workflow distributions of each procedure in~\ourdata, `Resecting' is omitted for better visualization.}
	\label{dataset_big}
\end{figure*}

\section{Related Work}
Surgical workflow recognition aims to improve the efficiency, safety, and outcomes of surgical procedures~\cite{deepphase,opera,tmrnet,sahc,Pelphix,last,lessismore,weaklysupervised}. This task involves the automatic identification and classification of various phases of surgery, providing real-time decision support for surgeons, enhancing postoperative analysis and evaluation, and optimizing the overall surgical workflow~\cite{tecno}. Early methods~\cite{svrcnet} employed Convolutional Neural Networks (CNNs) and Recurrent Neural Networks (RNNs), particularly Long Short-Term Memory (LSTM) ~\cite{lstm} networks, for temporal modeling and hierarchical feature learning. Among them, SV-RCNet~\cite{svrcnet} integrates CNN and RNN to explore complementary information from visual and temporal features learned from surgical videos. Endo3D~\cite{endo3d} utilizes the LSTM network to extract coarse-level information for online prediction. To address the slow training speed and limited receptive field of RNNs, TeCNO~\cite{tecno} performs hierarchical prediction refinement with causal and dilated MS-TCNs for fine-grained surgical phase recognition. Further, SWNet~\cite{swnet} implements deep 3D CNNs and utilizes prior knowledge noise filtering to improve MS-TCN.  

Recent studies~\cite{opera,transsvnet,skit,lovit} adopted transformers~\cite{attentionisallyouneed} to better model temporal relations in long sequences, enabling finer recognition and contextual understanding of surgical activities. For instance, OperA~\cite{opera} proposes a novel attention regularizer to achieve higher feature quality. Trans-SVNet~\cite{transsvnet} aggregates spatial and temporal embeddings for active queries with higher inference speed. Latest, Surgformer~\cite{surgformer} proposes a novel hierarchical temporal attention to capture both global and local information to enhance the overall temporal representation, demonstrating the potential of transformers in advancing surgical phase recognition.
For better temporal modeling of short-term and long-term information, in this paper, we exploit dense attention from the transformer combined with Mamba-based block~\cite{mamba,mamba2} in the proposed MTE and CSM, respectively, for effective surgical workflow recognition. 

\begin{table}[t]
\centering
\renewcommand{\arraystretch}{1.2}
\renewcommand{\tabcolsep}{3.6mm}
\begin{tabular}{rl}
\hline
Workflows & Descriptions \\
\hline
\textbf{Preparing} & \textit{Preparation stage.}  \\
\textbf{Knotting} & \textit{Knotting of the Foley catheter.}\\
\textbf{Resecting} & \textit{Procedure of the liver resection.}  \\
\textbf{Releasing} & \textit{Release of the Foley catheter.}  \\
\textbf{Postprocessing} & \textit{Postprocessing stage.} \\
\hline
\end{tabular}
\caption{\textbf{Workflow descriptions.} See visualization in~\figref{start}.}
\label{definition}
\end{table}

\section{PmLR50 Dataset}
Developing new tasks and datasets for surgical workflow recognition facilitates comprehensive intraoperative intelligent assistance and postoperative assessment, driving the development of surgical video analysis. 
According to the clinical needs of laparoscopic liver surgery, we contribute a new benchmark for liver resections with the Pringle maneuver, named~\ourdata. This benchmark embraces two challenging and complementary tasks: Pringle maneuver workflow recognition and blocking effectiveness detection. Exemplars of~\ourdata~are exhibited in~\figref{start}. We will provide detailed information of~\ourdata~in three key aspects: image collection, professional annotation, and dataset statistics.

\noindent\textbf{Image Collection:}
Data collection for surgical video analysis demands rigorous standards for dataset size and annotation quality. To ensure high-quality and representative data, we collect 25,037 high-resolution (1280$\times$720) video frames from 50 liver resection video clips containing Pringle maneuver procedures, which are carefully selected by six hepatobiliary surgeons at the collaborating hospital. Specifically, 35 clips are used for training, 5 for validation, and 10 for testing. The distribution of each surgical phase in~\ourdata~is shown in~\figref{dataset_big}. The frames from the Knotting and Releasing phases are sampled at 3 fps, while the frames from other phases are sampled at 0.33 fps to provide a more balanced data distribution, providing strong support for the development of workflow recognition frameworks. In addition, each surgical clip contains an average of 501 frames, with the shortest clip containing 313 frames and the longest clip containing 726 frames, ensuring that each phase accurately reflects the corresponding operation.

\begin{figure}[t]
	\centering
	\includegraphics[scale=.142]{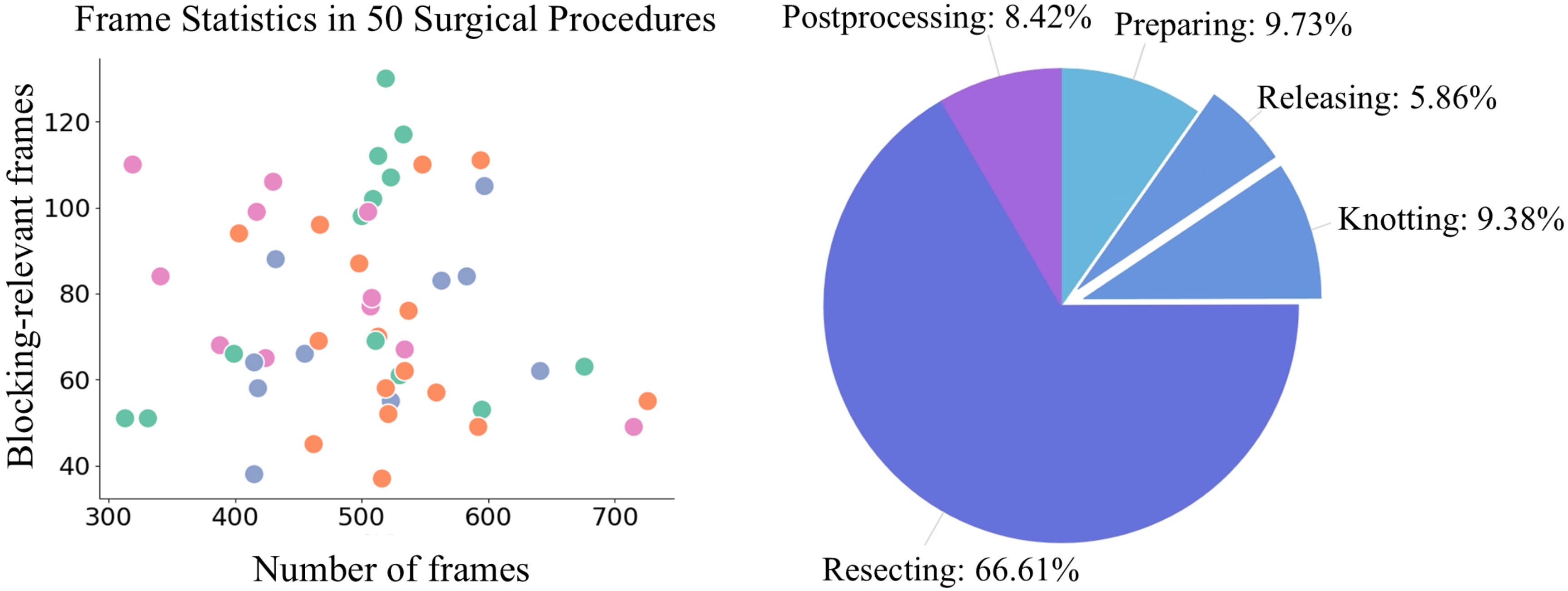}
	\caption{\textbf{Left:} The statistics of the number of blocking-relevant frames (Knotting and Releasing) with respect to the number of frames in each surgical procedure. \textbf{Right:} The distribution of the five surgical workflows in~\ourdata.}
	\label{pie}
\end{figure}

\begin{figure*}[t]
	\centering
	\includegraphics[scale=.354]{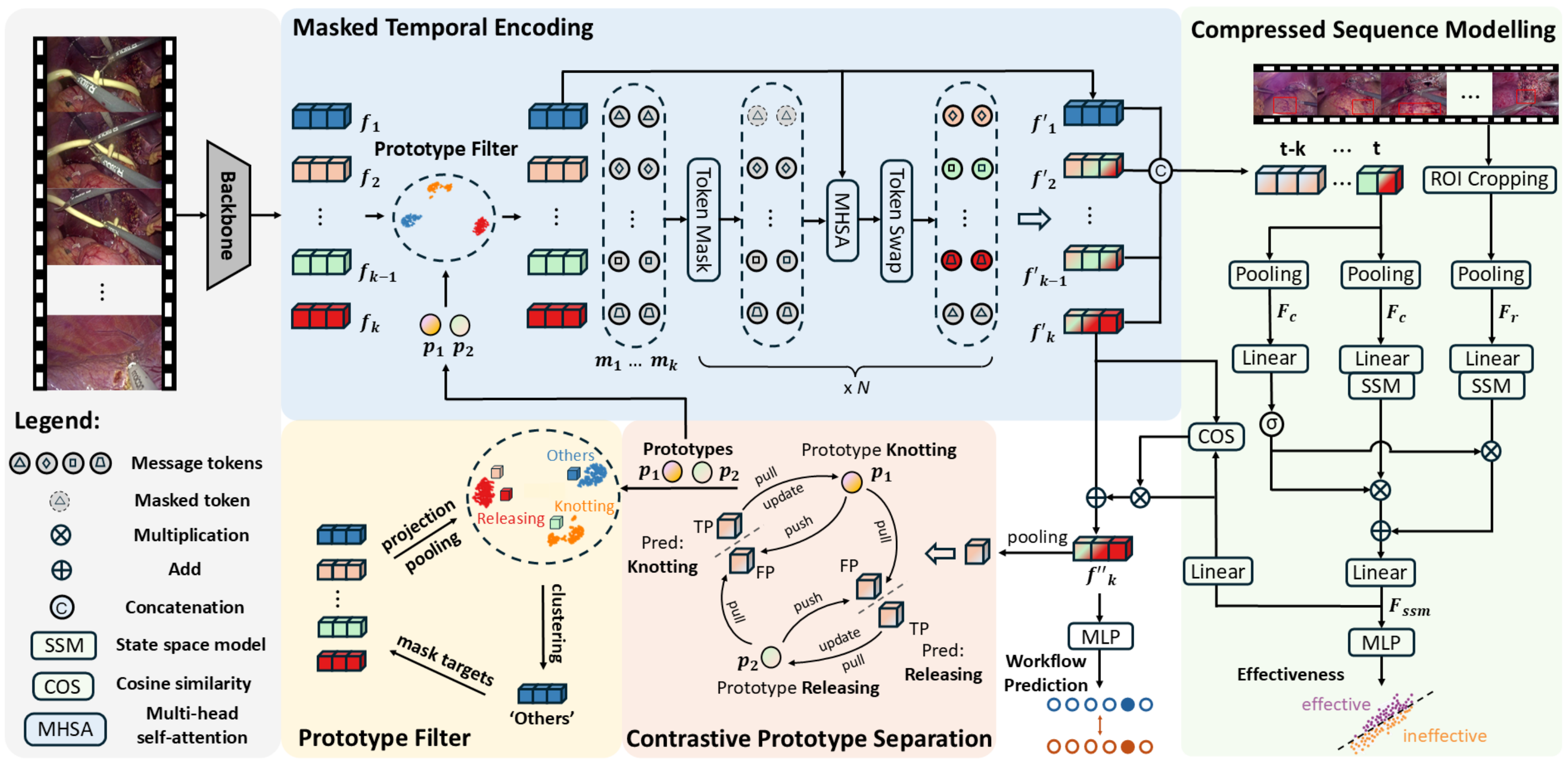}
	\caption{Overview of the proposed~\ourmodel~framework. Our model consists of a Masked Temporal Encoding (MTE), a Compressed Sequence Modeling (CSM), and a Contrastive Prototype Separation (CPS) for Pringle maneuver workflow recognition. CSM also conducts blocking effectiveness detection through efficient long-term memory modeling. Moreover, CPS employs a contrast learning strategy to enhance the ability to distinguish between confusing intraoperative activities in the feature space.}
	\label{main}
\end{figure*}

\noindent\textbf{Professional Annotation:}
When creating a large surgical dataset, data annotation is crucial for quality and reliability. To ensure the labeling quality of~\ourdata, we invite six experienced hepatobiliary surgeons from collaborating hospitals to elaborately annotate and review each video clip. During labeling, surgeons refer to static frames and dynamic videos to build our dataset from three aspects: surgical workflows, blocking effectiveness, and tissue ischemic regions.

\begin{itemize}
\item \noindent\textbf{Surgical Workflows}. For surgical workflow annotation, surgeons divide the procedure of PM into five phases: \emph{Preparing, Knotting, Resecting, Releasing, and Postprocessing}. Refer to~\tabref{definition} for detailed descriptions. 

\item \noindent\textbf{Blocking Effectiveness}. Beyond PM workflow recognition, surgeons also provide corresponding binary labels for video frames associated with blocking actions to determine the effectiveness. The ineffective blocking is defined as ``inability to successfully intermittently block blood inflow to the liver by knotting the catheter''~\cite{Sidaway2024}. In the~\ourdata~dataset, we provide a total of 5 surgical cases of ineffective blocking, representing 10\% of the total cases.

\item \noindent\textbf{Tissue Ischemia Regions}. To assist in the blocking effectiveness detection of PM, surgeons also provide box labels for liver ischemia regions. As shown in the bottom of~\figref{start}, the area in boxes represents the location where the most significant changes in liver ischemia states occur during surgery.
\end{itemize}

\noindent\textbf{Dataset Characteristics and Statistics:}
As depicted in the left of~\figref{pie}, the majority of the sampled video frames in~\ourdata~range between 400 and 600 frames. Among a total of five surgical phases, `Resecting' contains the highest number of frames, accounting for 66.61\% of the total frames, significantly higher than the 9.38\% for `Knotting' and 5.86\% for `Releasing' (see the right of~\figref{pie}). Notably, our dual-rate frame sampling (\ie, 3 fps and 0.33 fps) ensures more than 30 frames of blocking-related operations per surgical procedure, thus providing more balanced samples for various surgical phases.

\section{PmNet Method}
\subsection{Framework Overview}
\figref{main} illustrates our~\ourmodel~framework for PM workflow recognition and blocking effectiveness detection. Specifically, we first utilize EfficientNet-B3~\cite{efficientnet} as the backbone to extract frame-level features from video streams. Then, Masked Temporal Encoding (MTE) interacts with the visual features using swapped message tokens, performing efficient short-term temporal modeling. Here, blocking-irrelevant features are masked for effective dynamic aggregation with a continually updated prototype filter. Subsequently, we adopt Compressed Sequence Modeling (CSM) to establish long-range dependencies on temporally pooled long-term features and information within tissue regions for blocking effectiveness detection. Afterward, short-term memories interact with long-term memories for contextual retrieval, and then temporal-enhanced features pass through MLP for PM workflow recognition. Additionally, Contrastive Prototype Separation (CPS) is embedded to expand the feature space distance between target prototypes during training, thereby enhancing the discrimination capability for similar intraoperative actions. The detailed explanations of MTE, CSM, and CPS are provided below.

\subsection{Masked Temporal Encoding}
Fine-grained temporal information interaction is essential for capturing intraoperative surgical operations, as the detailed continuous motions between consecutive frames provide vital clues for distinguishing similar actions (\eg, Knotting and Releasing).
To this end, we introduce Masked Temporal Encoding (MTE), which leverages the dense attention mechanism from transformers combined with swapped message tokens for short-term temporal integration of fine-grained action details in video streaming. 
Given the feature sequence $F\in \mathbb{R}^{ N\times c}$ of $N$ consecutive frames from EfficientNet-B3~\cite{efficientnet}, we employ a temporal window size of $w$ to divide the sequence into $N/w$ non-overlapping clip features $f_k\in \mathbb{R}^{ w\times c}$, where $k\in N$ and $k\le N/w$.
For each clip, we initialize $d$ message tokens $m_k \in \mathbb{R}^{d \times c}$ and concatenate them with the corresponding clip features $f_k$, serving as Queries, Keys, and Values for intra-clip temporal integration:
\begin{equation}\label{Eq:gene_qkv}
\begin{aligned}
    Q &= FC(\mathcal{C}(f_k, m_k)),\ K = FC(\mathcal{C}(f_k, m_k)), \\
                     &\hspace{1.2cm} V = FC(\mathcal{C}(f_k, m_k)),
\end{aligned}
\end{equation}
where $FC(\cdot)$ and $\mathcal{C}(\cdot)$ represent fully connected layers and concatenation operation, respectively. 
Next, we use self-attention for temporal integration within clips to obtain temporal aggregated features ${f_k}'$.
After each self-attention operation, we perform a token-swap strategy to encourage inter-clip temporal aggregation across clips through swapped message tokens. Given $N/w$ clips from the input sequence, we perform $\lceil (\sqrt{8N - 7}-1)/2 \rceil $ swaps in total, \eg, the $i$-th swap is implemented as ${m}_{k} = {m}_{k-i}, {m}_{k-i} = {m}_{k}$. In this way, we can accomplish inter-clip temporal interactions with as few token swap operations as possible. 

Considering the large number of blocking-irrelevant frames in video streaming, we also employ a continually updated prototype filter to emphasize the surgical action details. Specifically, given temporally aggregated features ${f''_k}$ (refer to Eq. \ref{Eq:COS}) for true positive (TP) samples, we consider them to have better intra-class consistency. Those features are gathered and temporally pooled for a generalized temporal representation to update the prototype of the corresponding classes via an exponential moving average (EMA):
\begin{equation}\label{Eq:ema}
	p_j = (1 - \alpha) \cdot \frac{1}{n_{TP}^j} \sum_{i \in s_{TP}^j} \text{P}_t((f''_k)_i) + \alpha \cdot p_j,
\end{equation} 
where $p_j$ represents the feature prototype for the $j$-th surgical phase, $(f''_k)_i$ denotes the $i$-th clip feature ${f''_k}$ in the whole training procedure, $n^j$ means the number of TP samples within the TP set $s_{TP}^j$ for the $j$-th phase, $\alpha$ is the momentum term, and ${\text{P}_t}(\cdot)$ denotes the temporal pooling operation. After that, we calculate the minimum cosine similarity between the temporal features ${f''_k}$ from clips and each prototype to determine the operation relevance $R_{kj}$ of the $k$-th clip to prototype $p_j$:
\begin{equation}\label{Eq:contrast}
	R_{kj} = \frac{\text{P}_t(f''_k)\cdot p_j}{\|\text{P}_t(f''_k) \|\cdot \|p_j  \|}.
\end{equation} 
For blocking-related clip features not clustered as Knotting/Releasing according to the operation relevance $R_{kj}$, we masked their information tokens for more efficient temporal aggregation of surgical operation details.

\begin{figure*}[t]
	\centering
 \includegraphics[width=0.98\linewidth]{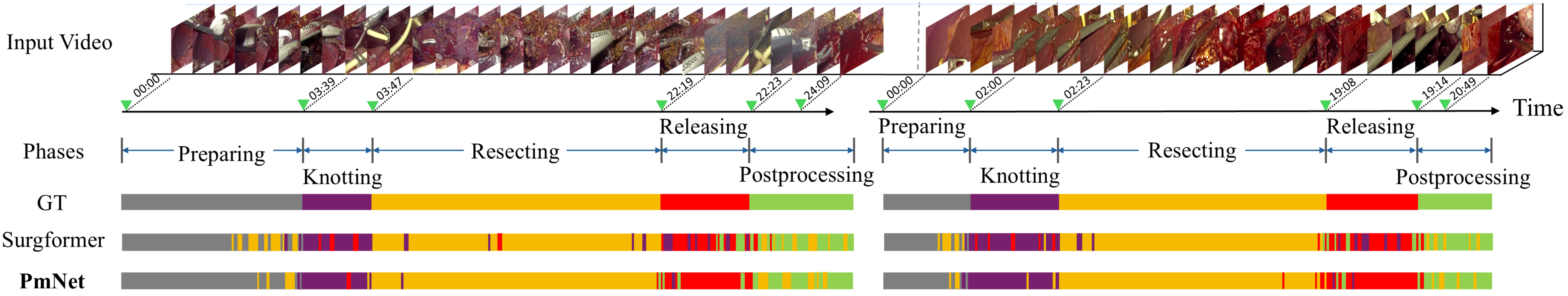}
	\caption{Visualization of PM workflow recognition predictions by color-coded ribbon on the~\ourdata~test set.}
	\label{vis}
\end{figure*}

\subsection{Compressed Sequence Modeling}
Long-term real-time monitoring of the liver ischemia and bleeding state is important for assessing the blocking effectiveness. The long-term temporal memory also contributes to understanding the whole surgical procedure and tracking the relative periods of surgical operations. However, when dealing with long sequences, the dense attention mechanism in transformer-based models imposes significant computational costs. To address this issue, we design the Compressed Sequence Modeling (CSM), which leverages the state space model (SSM) to perform compressed feature modeling by utilizing temporally pooled long-term features and liver ischemia region information. 

Unlike the vanilla Mamba block~\cite{mamba}, CSM embeds an overlapping pooling operation to efficiently interact with video sequences and incorporates a local perceptual branch to enhance low-level features of the target tissue. Given temporally encoded features $f'_k$ from MTE and cropped tissue region sequence $F_r$ from frames, we first concatenate $f'_k$ as long-term memory and then apply overlapping temporal pooling across the time dimension to obtain a consistent feature representation $F_c$:
\begin{equation}\label{Eq:pooling}
	F_c = {\text{P}_t}({\mathcal{C}}(\left \{ f'_k \right \})),\text{ } k = 1,2,..., N/w,
\end{equation}
where ${\text{P}_t}(\cdot)$ and ${\mathcal{C}}(\cdot)$ represent temporal pooling and concatenation, respectively. Afterward, we adopt Mamba-based operation~\cite{mamba,mamba2} for efficient long-term temporal modeling:
\begin{equation}\label{Eq:SSM}
F_{ssm} = \sigma(Ln(F_c))(\mathcal{S}(Ln(F_c)) + \mathcal{S}(Ln(F_r))),
\end{equation} 
where $\sigma(\cdot)$, $\mathcal{S}(\cdot)$, and $Ln(\cdot)$ denote the activation function, SSM operation, and linear projection respectively. Then, we use an MLP to predict the blocking effectiveness from the sequence. 
Furthermore, we use temporally aggregated features $f'_k$ as Queries, and the compressed long-term temporal representation $F_{ssm}$ from SSM as Keys and Values to facilitate contextual retrieval with cosine similarity:
\begin{equation}\label{Eq:COS}
f''_k = Softmax\left(\frac{f'_k F_{ssm}^T}{\|f'_k\|\cdot \|F_{ssm}\|}\right)F_{ssm} + f'_k,
\end{equation} 
where $f''_k$ denotes the clip features embedded with the long-term memory $F_{ssm}$. 
Finally, $f''_k$ integrates with the short-term aggregated $f'_k$ and long-term temporal memory $F_{ssm}$ for surgical workflow recognition through MTE and CSM.

\begin{table*}[t!]
\centering
\footnotesize
\renewcommand{\arraystretch}{1.1}
\setlength\tabcolsep{8pt}
\begin{tabular}{c|c|c|c|c|c|c}
\hline
 Tasks & Metrics & TeCNO & TMRNet & Trans-SVNet & Surgformer & \textbf{\ourmodel} \\ \hline 
 \multirow{4}{*}{\shortstack{PM Workflow\\Recognition}} 
  & Precision $\uparrow$ & 88.49  &82.78 & 88.10 &84.58 &\textbf{92.36} \\
  & Recall $\uparrow$    & 84.52  &70.78 & 86.32 &61.45& \textbf{86.93} \\
  & Accuracy $\uparrow$  & 92.53  &86.59 & 93.08 &84.51& \textbf{93.83} \\
  & Jaccard  $\uparrow$  & 75.12  &61.66 & 79.10 &57.98 & \textbf{81.47} \\
  \hline 
 \multirow{4}{*}{\shortstack{Blocking Effectiveness\\Detection}} 
   & Precision $\uparrow$    & 53.48 & 54.15  & 64.00  &53.13  & \textbf{65.33} \\
  &  Recall $\uparrow$       & 68.97 & 58.28 & 71.72 &63.45  &  \textbf{84.48}\\ 
    & Accuracy $\uparrow$    & 97.27 & 97.29 &  97.96  & 97.25 &  \textbf{98.21} \\
  & Jaccard $\uparrow$       & 43.11 & 39.02  & 51.10  &40.68& \textbf{58.33} \\ \hline 
   \multicolumn{2}{c|}{Params $\downarrow$} &  \textbf{24.74M}  & 63.12M  &  24.75M &  121.26M  & 66.62M \\  \hline
 \multicolumn{2}{c|}{Inference Speed (FPS) $\uparrow$} & - &  - &  -  &  29.85  &  \textbf{35.34} \\  \hline
\end{tabular}
\caption{Quantitative comparison with the previous state-of-the-art methods for PM workflow recognition and blocking effectiveness detection on the~\ourdata~test set. `-' indicates that the inference speed of two-stage models is omitted.}
\label{results_1}
\end{table*}

\subsection{Contrastive Prototype Separation}
Compared to other surgical procedures, liver resections with PM involve two easily confused phases: Knotting and Releasing. The visual characteristics of these two operations are highly similar, making it difficult for the model to accurately distinguish. To address this challenge, we propose a Contrastive Prototype Separation (CPS) strategy, which leverages the concept of contrastive learning to distinguish temporal prototypes of similar surgical operations in the feature space. 
We assume that each sample should be closer to the corresponding prototype while far away from other samples. The distance between two feature vectors, $\text{dis}(x, y)$, is defined by Euclidean distance:
\begin{equation}\label{Eq:dis}
	\text{dis}(\bm{u}, \bm{v}) = \|\bm{u} - \bm{v}\|,
\end{equation}
where $\bm{u}$ and $\bm{v}$ stand for 1-D vectors. 
To discover the ambiguous samples promptly during the training stage, we introduce a contrastive loss $\mathcal{L}_{CL}$ to calibrate the temporal features $f''_k$ of those false positive (FP) samples:
\begin{equation}\label{Eq:contrast2}
	\mathcal{L}_{CL} = \frac{1}{2} \|f''_k - p_y\|^2 + \frac{1}{2} \max(0, 1 - \|f''_k - p_{\hat{y}}\|)^2,
\end{equation}
where $p_y$ and $p_{\hat{y}}$ are feature prototypes of the correct/predicted class, respectively. 
By minimizing the contrastive loss $\mathcal{L}_{CL}$, FP samples progressively move closer to the correct prototype and further away from the incorrect category prototype, thus facilitating discrimination between similar intraoperative activities. 

During training, the objective function of surgical workflow recognition $\mathcal{L}^s_{CE}$ and blocking effectiveness detection $\mathcal{L}^e_{CE}$ is measured via the cross-entropy loss:
\begin{equation}\label{eq:L_CE}
\mathcal{L}_{CE} = - \frac{1}{N} \sum_i 	\sum_j y_{ij} \mbox{log}(p_{ij}),
\end{equation}
where $N$ denotes the batch size. $y_{ij}$ is the one-hot label of the $i$-th sample, with $y_{ij} = 1$ if and only if $j$ is the target class. $p_{ij}$ is the prediction score of the $i$-th sample for class $j$.
Finally, the overall loss function of our model  $\mathcal{L}_{total}$ integrates the cross-entropy loss ($\mathcal{L}^s_{CE}$ and $\mathcal{L}^e_{CE}$) and contrastive loss $\mathcal{L}_{CL}$ in Eq. \ref{Eq:contrast}: 
\begin{equation}\label{Eq:final}
    \mathcal{L}_{total} = \mathcal{L}^s_{CE} + \mathcal{L}^e_{CE} + \lambda_{cl}\mathcal{L}_{CL},
\end{equation}
where the loss weight $\lambda_{cl}$ for $\mathcal{L}_{CL}$ is empirically set to 0.1. 

\section{Experiments}
\subsection{Experimental Settings}

\textbf{Implementation Details:}
Our model is trained with the AdamW optimizer for 50 epochs on two NVIDIA 4090 GPUs. 
The batch size is set to 16, and the initial learning rate is 3e-5. 
We adopt EfficientNet-B3~\cite{efficientnet} pre-trained on ImageNet~\cite{imagenet} as the backbone.
The parameters of other components in~\ourmodel~are randomly initialized.
In our experiments, 20 consecutive frames with frame rate $R=8$ are selected as model inputs in an online manner, with the clip window size $w$ set to 4. During training, we apply color jitter and random horizontal flip for data augmentation.
To trade off the accuracy and efficiency of~\ourmodel, we embed four token swap operations in MTE, with the number of SSM blocks set to 2 in CSM.

\noindent\textbf{Datasets and Evaluation Metrics:}
We adopt the proposed~\ourdata~dataset, which comprises 50 cases of laparoscopic liver resections with PM, to conduct surgical workflow recognition and blocking effectiveness detection. All comparative experiments are conducted on the test set containing ten PM procedures, while ablation studies are performed on the validation set containing five cases. Following previous research~\cite{transsvnet,tecno,surgformer}, we adopt four metrics for evaluation: Precision, Recall, Accuracy, and Jaccard index. Additionally, we evaluate the inference speed of the model on one single NVIDIA 4090 GPU. Notably, for the assessment of blocking effectiveness detection, we focused solely on frames from the `Knotting' phase in surgery procedures, which are labeled with the category `valid' or `invalid'.

\subsection{Comparison with State-of-the-art Methods}
We comprehensively compare the experimental results of~\ourmodel~with four state-of-the-art models for~\cite{tecno,tmrnet,transsvnet,surgformer} PM workflow recognition and blocking effectiveness detection.
All compared methods are implemented using official code, and an extra detection head is added to enable blocking effectiveness detection in line with~\ourmodel. 

As illustrated in~\tabref{results_1}, our model shows superior performance in both tasks. Compared to the second-best model Trans-SVNet, our model achieves an improvement of 2.37\% Jaccard in PM workflow recognition. Besides, by modeling long-term memory through CSM, \ourmodel~obtains a high accuracy of 98.21\% in blocking effectiveness detection. In addition, our model offers certain advantages in terms of parameters and inference efficiency, reaching an inference speed of 35.34 fps for real-time clinical applications. \ourmodel~also outperforms other methods across all phases, which can be attributed to CPS to discriminate between confused intraoperative actions. 
\figref{vis} visualizes the predictions of our method compared to the latest Surgformer for PM workflow recognition, where~\ourmodel~demonstrates better consistency and accuracy across all surgical phases. 
 
\subsection{Ablation Studies}
\noindent\textbf{Ablations for Each Component.} 
\tabref{table_component} ablates the contribution of each component of~\ourmodel~to surgical workflow recognition. The experimental results indicate that all components, including MTE, CSM, and CPS play a positive role. The results in the last two rows illustrate that MTE used for capturing short-term temporal information shows the most significant effect, increasing the Jaccard value for PM workflow recognition by 6.44\%. Meanwhile, the efficient long-term temporal modeling by CSM and the comparative learning of intraoperative activities by CPS also enhance the validity and robustness of~\ourmodel.

\begin{figure}[t!]
	\centering
 \includegraphics[width=0.74\linewidth]{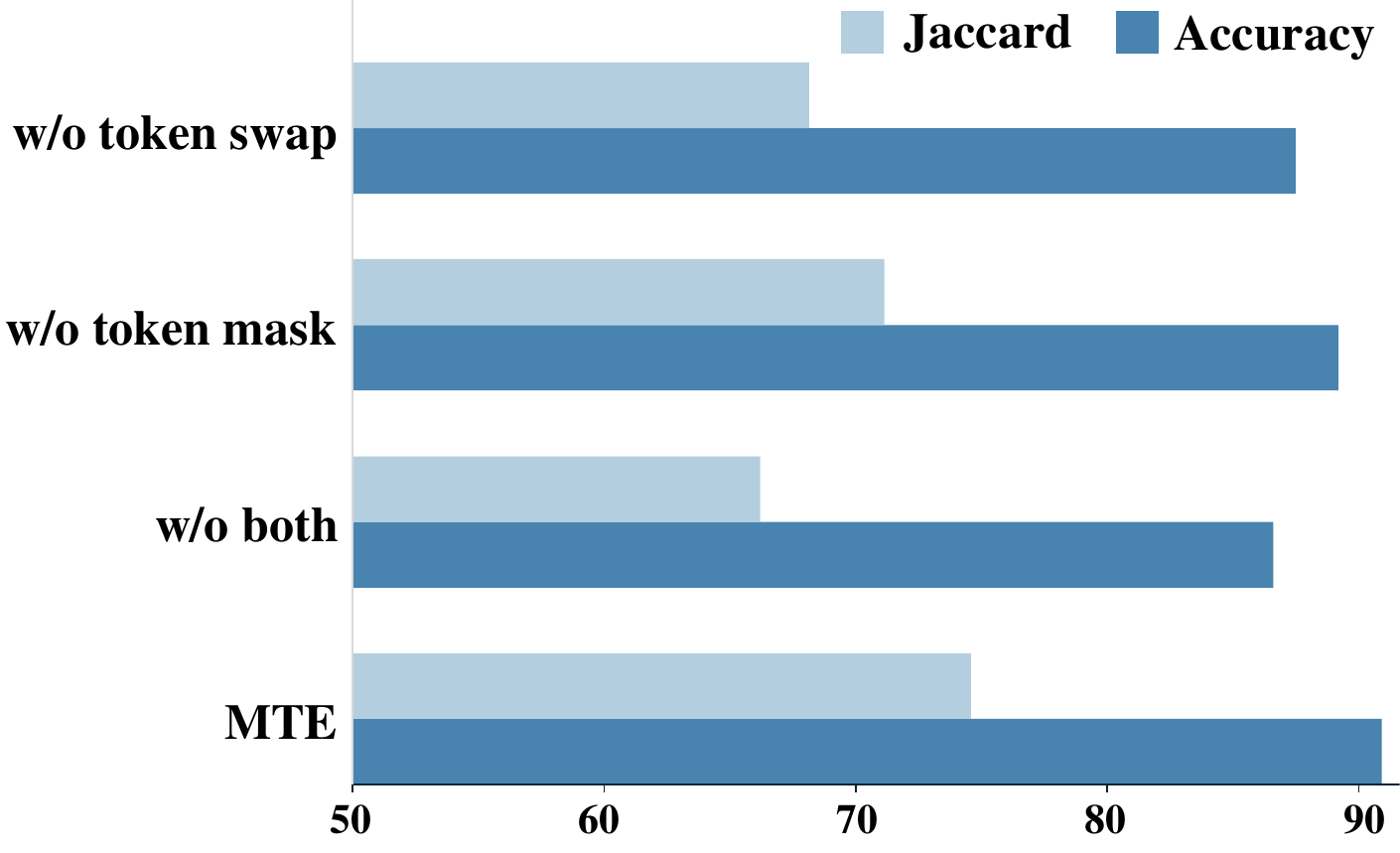}
	\caption{Ablations for the effect of MTE on PM workflow recognition in blocking-relevant phases.}
	\label{ablation_mte}
\end{figure}

\begin{table}[t!]
\centering
\footnotesize
  \renewcommand{\arraystretch}{1.0}
  \setlength\tabcolsep{2.9pt}
\begin{tabular}{c|c|c|c|c|c|c}
\hline
MTE & CSM & CPS  & Precision $\uparrow$ & Recall $\uparrow$ & Accuracy $\uparrow$& Jaccard $\uparrow$\\ \hline
\cmark  & \cmark & \xmark  & \textbf{90.83}  & 79.11 & 89.57 & 73.04 \\ 
\cmark  &  \xmark  & \cmark & 88.16 & 79.25 & 89.18 & 71.13\\ 
\xmark  & \cmark & \cmark & 85.19 & 76.85 & 87.49 & 68.14\\ 
\rowcolor[RGB]{220,238,246}
\cmark  & \cmark & \cmark  & 90.61 & \textbf{80.52} & \textbf{90.91} & \textbf{74.58}\\  

\end{tabular}
\caption{Contributions of each component of~\ourmodel~for PM workflow recognition on the~\ourdata~validation set.}
\label{table_component}
\end{table}

\begin{table}[t!]
\centering
\footnotesize
  \renewcommand{\arraystretch}{1.}
  \setlength\tabcolsep{3.5pt}
\begin{tabular}{c|c|c|c|c|c}
\hline
  Phases  & Preparing & Knotting & Resecting & Releasing & Post. \\ \hline
  Preparing       &  -  & 73.11 & 72.58 &72.66 &73.58\\
  Knotting        &  73.11  & - & 72.06  & \textbf{74.58} &72.82\\
  Resecting       &  72.58  & 72.06 & - &72.26 &73.55 \\
  Releasing       &  72.66  & \textbf{74.58} & 72.26 &- &71.68\\
  Post.  &  73.58  &72.82 & 73.55 &71.68 &-\\
  
\end{tabular}
\caption{Effect of CPS on Jaccard for PM workflow recognition. `Post.' denotes the Postprocessing phase.}
\label{ablation_cps}
\end{table}

\noindent\textbf{Effect of MTE.}
We investigate the influence of the token swap operation and the masking mechanism in MTE. 
As shown in \figref{ablation_mte}, the token swap operation greatly improves the performance in the blocking-related phase, which demonstrates the effectiveness of our inter-clip temporal interaction in capturing short-term actions. Besides, the masking mechanism filters out blocking-irrelevant features and effectively aggregates short-term temporal dynamics.
\begin{figure}[t!]
\centering
\includegraphics[width=0.87\linewidth]{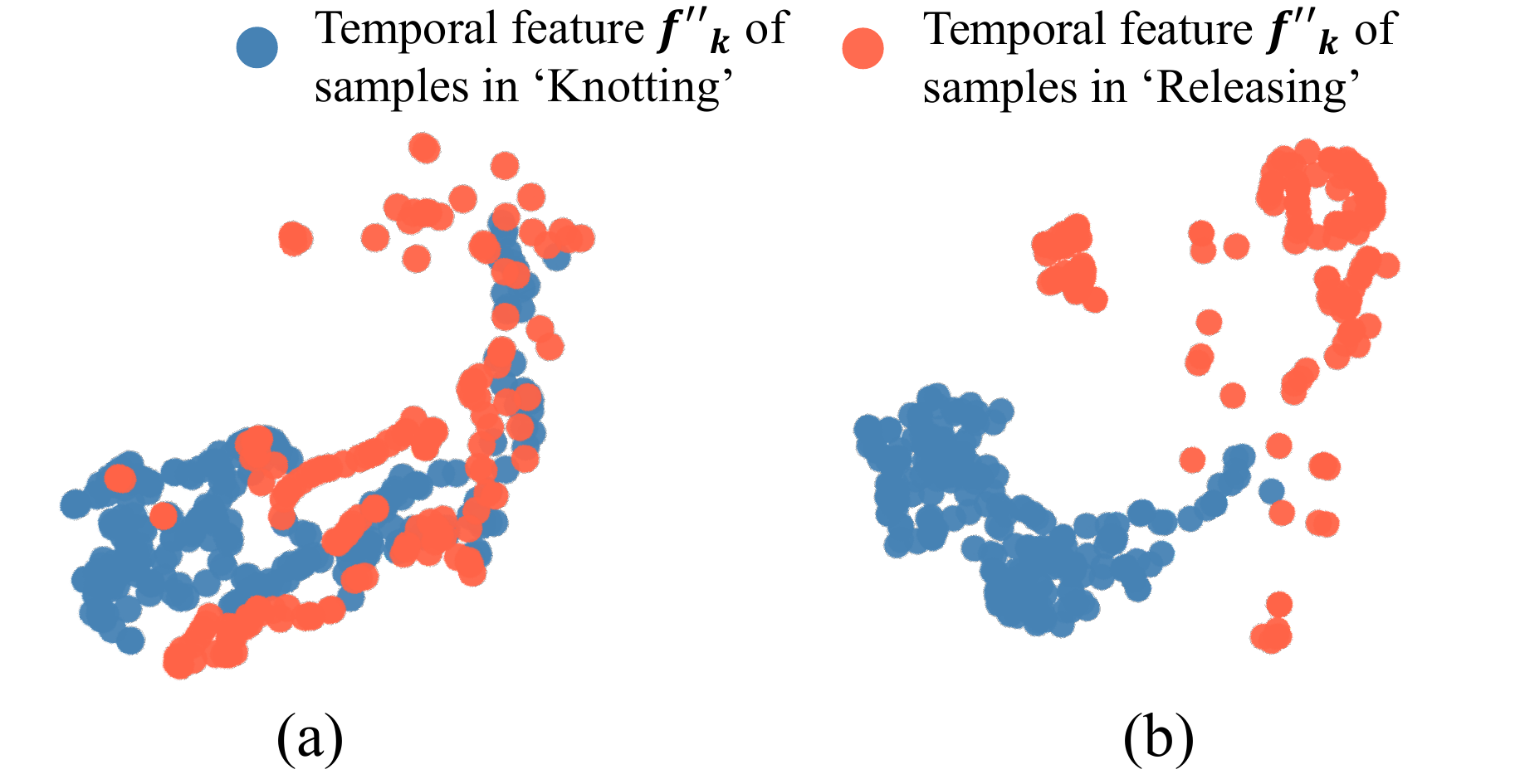}
\caption{(a) t-SNE of blocking-relevant features without CPS. (b) t-SNE of blocking-relevant features with CPS.}
\label{tsne}
\end{figure}

\noindent\textbf{Effect of CSM.}
We discuss the effect of the design in CSM on block effectiveness detection. The results in \tabref{ablation_csm} show that the tissue region information is effective for long-term monitoring of liver ischemic states. Temporal pooling operation and SSM blocks also contribute to enhancing performance, thanks to efficient long-term memory modeling.

\begin{table}[t!]
\centering
\footnotesize
  \renewcommand{\arraystretch}{1.}
  \setlength\tabcolsep{3.6pt}
\begin{tabular}{r|c|c|c|c}
\hline
 Configs  & Precision $\uparrow$ & Recall $\uparrow$ & Accuracy $\uparrow$  & Jaccard $\uparrow$ \\ \hline
  w/o Pooling  &  64.10  & \textbf{86.21} & 98.28 & 58.14\\
  w/o Box label & \textbf{82.76} & 46.15 & 98.24 &42.11\\
  w/o SSM  & 40.62  & 74.29 & 96.28 & 35.63 \\
  \rowcolor[RGB]{220,238,246}
  CSM  &71.79 & 80.00 & \textbf{98.57} & \textbf{60.87}\\

\end{tabular}
\caption{Effect of CSM for blocking effectiveness detection.}
\label{ablation_csm}
\end{table}

\noindent\textbf{Contrastive Pairs in CPS.}
We compare the performance of~\ourmodel~with different contrastive learning pairs in~\tabref{ablation_cps}. 
Interestingly, apart from the `Knotting-Releasing' pair adopted in CPS, other contrastive pairs with low Jaccard index contribute little to  workflow recognition. 
This is probably attributed to the different complexity levels of these phases. 
Further, \figref{tsne} shows the t-SNE results of blocking-relevant features in the latent space from~\ourmodel, where the points represent temporal features $f''_k$. Samples with the same phase are denoted with the same color. we can observe that points of different colors from `Knotting' and `Releasing' phases are clearly distinguished, indicating that CPS is effective in distinguishing similar intraoperative actions.

\section{Conclusion}
This paper suggests two AI-assisted surgical video analysis tasks geared towards liver resections with the Pringle maneuver: workflow recognition and blocking effectiveness detection. Accordingly, we establish a comprehensive benchmark, \ourdata, to advance research in laparoscopic liver surgery. 
Meanwhile, we propose a unified online baseline for both tasks, termed \ourmodel, which efficiently models short-term action details and long-term temporal memory with Masked Temporal Encoding (MTE) and Compressed Sequence Modeling (CSM). Contrastive Prototype Separation (CPS) is also embedded to improve the discrimination between similar intraoperative actions. 
Experimental results on~\ourdata~demonstrate that~\ourmodel~achieves superior performance in both PM workflow recognition and blocking effectiveness detection. 
This fundamental research has great potential to be applied to practical laparoscopic liver surgery.

\section{Acknowledgments}
The work described in this paper was supported in part by the Research Grants Council of the Hong Kong Special Administrative Region, China, under Project T45-401/22-N; and  
by the Regional Joint Fund of Guangdong (Guangdong–Hong Kong–Macao Research Team Project) under Grant 2021B1515130003; and by the Shenzhen Science and Technology Program, under Grant RCYX20231211090127030.
\bibliography{aaai25}

\begin{thebibliography}{33}
\providecommand{\natexlab}[1]{#1}

\bibitem[{Cao et~al.(2023)Cao, Yip, Chen, Scheppach, Luo, Yang, Cheng, Long, Jin, Chiu, Yam, Meng, and Dou}]{Cao2023}
Cao, J.; Yip, H.-C.; Chen, Y.; Scheppach, M.; Luo, X.; Yang, H.; Cheng, M.~K.; Long, Y.; Jin, Y.; Chiu, P. W.~Y.; Yam, Y.; Meng, H. M.-L.; and Dou, Q. 2023.
\newblock Intelligent surgical workflow recognition for endoscopic submucosal dissection with real-time animal study.
\newblock \emph{Nature Communications}, 14(1): 6676.

\bibitem[{Chen et~al.(2018)Chen, Feng, Lu, and Zhou}]{endo3d}
Chen, W.; Feng, J.; Lu, J.; and Zhou, J. 2018.
\newblock Endo3D: Online Workflow Analysis for Endoscopic Surgeries Based on 3D CNN and LSTM.
\newblock In \emph{MICCAI Workshop}, 97--107.

\bibitem[{Czempiel et~al.(2020)Czempiel, Paschali, Keicher, Simson, Feussner, Kim, and Navab}]{tecno}
Czempiel, T.; Paschali, M.; Keicher, M.; Simson, W.; Feussner, H.; Kim, S.~T.; and Navab, N. 2020.
\newblock TeCNO: Surgical Phase Recognition with Multi-stage Temporal Convolutional Networks.
\newblock In \emph{MICCAI}, 343--352.

\bibitem[{Czempiel et~al.(2021)Czempiel, Paschali, Ostler, Kim, Busam, and Navab}]{opera}
Czempiel, T.; Paschali, M.; Ostler, D.; Kim, S.~T.; Busam, B.; and Navab, N. 2021.
\newblock OperA: Attention-Regularized Transformers for Surgical Phase Recognition.
\newblock In \emph{MICCAI}, 604--614.

\bibitem[{Dao and Gu(2024)}]{mamba2}
Dao, T.; and Gu, A. 2024.
\newblock Transformers are {SSM}s: Generalized Models and Efficient Algorithms Through Structured State Space Duality.
\newblock In \emph{ICML}.

\bibitem[{Deng et~al.(2009)Deng, Dong, Socher, Li, Li, and Fei-Fei}]{imagenet}
Deng, J.; Dong, W.; Socher, R.; Li, L.-J.; Li, K.; and Fei-Fei, L. 2009.
\newblock ImageNet: a large-scale hierarchical image database.
\newblock In \emph{IEEE CVPR}, 248--255.

\bibitem[{Ding and Li(2022)}]{sahc}
Ding, X.; and Li, X. 2022.
\newblock Exploring Segment-Level Semantics for Online Phase Recognition From Surgical Videos.
\newblock \emph{IEEE TMI}, 41(11): 3309--3319.

\bibitem[{Ding et~al.(2023)Ding, Yan, Wang, Zhao, Zhuang, Xu, and Li}]{lessismore}
Ding, X.; Yan, X.; Wang, Z.; Zhao, W.; Zhuang, J.; Xu, X.; and Li, X. 2023.
\newblock Less Is More: Surgical Phase Recognition From Timestamp Supervision.
\newblock \emph{IEEE TMI}, 42(6): 1897--1910.

\bibitem[{Gu and Dao(2023)}]{mamba}
Gu, A.; and Dao, T. 2023.
\newblock Mamba: Linear-Time Sequence Modeling with Selective State Spaces.
\newblock \emph{arXiv preprint arXiv:2312.00752}.

\bibitem[{Hochreiter and Schmidhuber(1997)}]{lstm}
Hochreiter, S.; and Schmidhuber, J. 1997.
\newblock Long Short-Term Memory.
\newblock \emph{Neural Computation}, 9(8): 1735--1780.

\bibitem[{Jin et~al.(2018)Jin, Dou, Chen, Yu, Qin, Fu, and Heng}]{svrcnet}
Jin, Y.; Dou, Q.; Chen, H.; Yu, L.; Qin, J.; Fu, C.-W.; and Heng, P.-A. 2018.
\newblock SV-RCNet: Workflow Recognition From Surgical Videos Using Recurrent Convolutional Network.
\newblock \emph{IEEE TMI}, 37(5): 1114--1126.

\bibitem[{Jin et~al.(2021)Jin, Long, Chen, Zhao, Dou, and Heng}]{tmrnet}
Jin, Y.; Long, Y.; Chen, C.; Zhao, Z.; Dou, Q.; and Heng, P.-A. 2021.
\newblock Temporal Memory Relation Network for Workflow Recognition From Surgical Video.
\newblock \emph{IEEE TMI}, 40(7): 1911--1923.

\bibitem[{Jin et~al.(2022)Jin, Long, Gao, Stoyanov, Dou, and Heng}]{transsvnet}
Jin, Y.; Long, Y.; Gao, X.; Stoyanov, D.; Dou, Q.; and Heng, P.-A. 2022.
\newblock Trans-SVNet: hybrid embedding aggregation Transformer for surgical workflow analysis.
\newblock \emph{IJCARS}, 17(12): 2193--2202.

\bibitem[{Khajeh et~al.(2021)Khajeh, Shafiei, Al-Saegh, Ramouz, Hammad, Ghamarnejad, Al-Saeedi, Rahbari, Reissfelder, Mehrabi, Probst, and Oweira}]{Khajeh2021}
Khajeh, E.; Shafiei, S.; Al-Saegh, S. A.-H.; Ramouz, A.; Hammad, A.; Ghamarnejad, O.; Al-Saeedi, M.; Rahbari, N.; Reissfelder, C.; Mehrabi, A.; Probst, P.; and Oweira, H. 2021.
\newblock Meta-analysis of the effect of the pringle maneuver on long-term oncological outcomes following liver resection.
\newblock \emph{Scientific Reports}, 11(1): 3279.

\bibitem[{Killeen et~al.(2023)Killeen, Zhang, Mangulabnan, Armand, Taylor, Osgood, and Unberath}]{Pelphix}
Killeen, B.~D.; Zhang, H.; Mangulabnan, J.; Armand, M.; Taylor, R.~H.; Osgood, G.; and Unberath, M. 2023.
\newblock Pelphix: Surgical Phase Recognition fromX-Ray Images inPercutaneous Pelvic Fixation.
\newblock In \emph{MICCAI}, 133--143.

\bibitem[{Liu et~al.(2023{\natexlab{a}})Liu, Boels, Garc{\'i}a-Peraza-Herrera, Vercauteren, Dasgupta, Granados, and Ourselin}]{lovit}
Liu, Y.; Boels, M.; Garc{\'i}a-Peraza-Herrera, L.~C.; Vercauteren, T. K.~M.; Dasgupta, P.; Granados, A.; and Ourselin, S. 2023{\natexlab{a}}.
\newblock LoViT: Long Video Transformer for Surgical Phase Recognition.
\newblock \emph{ArXiv}, abs/2305.08989.

\bibitem[{Liu et~al.(2023{\natexlab{b}})Liu, Huo, Peng, Sparks, Dasgupta, Granados, and Ourselin}]{skit}
Liu, Y.; Huo, J.; Peng, J.; Sparks, R.; Dasgupta, P.; Granados, A.; and Ourselin, S. 2023{\natexlab{b}}.
\newblock SKiT: a Fast Key Information Video Transformer for Online Surgical Phase Recognition.
\newblock In \emph{IEEE ICCV}, 21074--21084.

\bibitem[{Lu et~al.(2024)Lu, Xu, Zheng, Li, Xie, Wang, Lin, Chen, Cao, Lin, Tu, Huang, Lin, Yao, Zheng, and Huang}]{Lu2024}
Lu, J.; Xu, B.-b.; Zheng, H.-L.; Li, P.; Xie, J.-w.; Wang, J.-b.; Lin, J.-x.; Chen, Q.-y.; Cao, L.-l.; Lin, M.; Tu, R.-h.; Huang, Z.-n.; Lin, J.-l.; Yao, Z.-h.; Zheng, C.-H.; and Huang, C.-M. 2024.
\newblock Robotic versus laparoscopic distal gastrectomy for resectable gastric cancer: a randomized phase 2 trial.
\newblock \emph{Nature Communications}, 15(1): 4668.

\bibitem[{Man et~al.(1997)Man, Fan, Ng, Lo, Liu, and Wong}]{man1997prospective}
Man, K.; Fan, S.-T.; Ng, I.~O.; Lo, C.-M.; Liu, C.-L.; and Wong, J. 1997.
\newblock Prospective evaluation of Pringle maneuver in hepatectomy for liver tumors by a randomized study.
\newblock \emph{Annals of surgery}, 226(6): 704--713.

\bibitem[{Okamura et~al.(2019)Okamura, Yamamoto, Sugiura, Ito, Ashida, Ohgi, and Uesaka}]{Okamura2019}
Okamura, Y.; Yamamoto, Y.; Sugiura, T.; Ito, T.; Ashida, R.; Ohgi, K.; and Uesaka, K. 2019.
\newblock Novel patient risk factors and validation of a difficulty scoring system in laparoscopic repeat hepatectomy.
\newblock \emph{Scientific Reports}, 9(1): 17653.

\bibitem[{Ramesh et~al.(2023)Ramesh, DalľAlba, Gonzalez, Yu, Mascagni, Mutter, Marescaux, Fiorini, and Padoy}]{weaklysupervised}
Ramesh, S.; DalľAlba, D.; Gonzalez, C.; Yu, T.; Mascagni, P.; Mutter, D.; Marescaux, J.; Fiorini, P.; and Padoy, N. 2023.
\newblock Weakly Supervised Temporal Convolutional Networks for Fine-Grained Surgical Activity Recognition.
\newblock \emph{IEEE TMI}, 42(9): 2592--2602.

\bibitem[{Schoeffmann et~al.(2018)Schoeffmann, Taschwer, Sarny, M{\"{u}}nzer, Primus, and Putzgruber}]{cataract101}
Schoeffmann, K.; Taschwer, M.; Sarny, S.; M{\"{u}}nzer, B.; Primus, M.~J.; and Putzgruber, D. 2018.
\newblock Cataract-101: video dataset of 101 cataract surgeries.
\newblock In \emph{ACM Multimedia Systems}, 421--425.

\bibitem[{Seibold et~al.(2022)Seibold, Hoch, Farshad, Navab, and F{\"u}rnstahl}]{thadataset}
Seibold, M.; Hoch, A.; Farshad, M.; Navab, N.; and F{\"u}rnstahl, P. 2022.
\newblock Conditional Generative Data Augmentation for Clinical Audio Datasets.
\newblock In \emph{MICCAI}, 345--354.

\bibitem[{Sidaway(2024)}]{Sidaway2024}
Sidaway, P. 2024.
\newblock Laparoscopic hemihepatectomy is safe and effective.
\newblock \emph{Nature Reviews Clinical Oncology}, 21(7): 484--484.

\bibitem[{Tan and Le(2021)}]{efficientnet}
Tan, M.; and Le, Q.~V. 2021.
\newblock EfficientNetV2: Smaller Models and Faster Training.
\newblock In \emph{ICML}.

\bibitem[{Tao, Zou, and Zheng(2023)}]{last}
Tao, R.; Zou, X.; and Zheng, G. 2023.
\newblock LAST: LAtent Space-Constrained Transformers for Automatic Surgical Phase Recognition and Tool Presence Detection.
\newblock \emph{IEEE TMI}, 42(11): 3256--3268.

\bibitem[{Twinanda et~al.(2017)Twinanda, Shehata, Mutter, Marescaux, de~Mathelin, and Padoy}]{micai16}
Twinanda, A.~P.; Shehata, S.; Mutter, D.; Marescaux, J.; de~Mathelin, M.; and Padoy, N. 2017.
\newblock EndoNet: A Deep Architecture for Recognition Tasks on Laparoscopic Videos.
\newblock \emph{IEEE TMI}, 36(1): 86--97.

\bibitem[{Vaswani et~al.(2017)Vaswani, Shazeer, Parmar, Uszkoreit, Jones, Gomez, Kaiser, and Polosukhin}]{attentionisallyouneed}
Vaswani, A.; Shazeer, N.; Parmar, N.; Uszkoreit, J.; Jones, L.; Gomez, A.~N.; Kaiser, L.; and Polosukhin, I. 2017.
\newblock Attention is all you need.
\newblock In \emph{NeurIPS}.

\bibitem[{Wagner et~al.(2023)Wagner, Müller-Stich, Kisilenko, Tran, Heger, Mündermann, Lubotsky, Müller, Davitashvili, Capek, Reinke, Reid, Yu, Vardazaryan, Nwoye, Padoy, Liu, Lee, Disch, Meine, Xia, Jia, Kondo, Reiter, Jin, Long, Jiang, Dou, Heng, Twick, Kirtac, Hosgor, Bolmgren, Stenzel, {von Siemens}, Zhao, Ge, Sun, Xie, Guo, Liu, Kenngott, Nickel, von Frankenberg, Mathis-Ullrich, Kopp-Schneider, Maier-Hein, Speidel, and Bodenstedt}]{heichole}
Wagner, M.; Müller-Stich, B.-P.; Kisilenko, A.; Tran, D.; Heger, P.; Mündermann, L.; Lubotsky, D.~M.; Müller, B.; Davitashvili, T.; Capek, M.; Reinke, A.; Reid, C.; Yu, T.; Vardazaryan, A.; Nwoye, C.~I.; Padoy, N.; Liu, X.; Lee, E.-J.; Disch, C.; Meine, H.; Xia, T.; Jia, F.; Kondo, S.; Reiter, W.; Jin, Y.; Long, Y.; Jiang, M.; Dou, Q.; Heng, P.~A.; Twick, I.; Kirtac, K.; Hosgor, E.; Bolmgren, J.~L.; Stenzel, M.; {von Siemens}, B.; Zhao, L.; Ge, Z.; Sun, H.; Xie, D.; Guo, M.; Liu, D.; Kenngott, H.~G.; Nickel, F.; von Frankenberg, M.; Mathis-Ullrich, F.; Kopp-Schneider, A.; Maier-Hein, L.; Speidel, S.; and Bodenstedt, S. 2023.
\newblock Comparative validation of machine learning algorithms for surgical workflow and skill analysis with the HeiChole benchmark.
\newblock \emph{Medical Image Analysis}, 86: 102770.

\bibitem[{Wang et~al.(2022)Wang, Lu, Long, Zhong, Cheung, Dou, and Liu}]{autolaparo}
Wang, Z.; Lu, B.; Long, Y.; Zhong, F.; Cheung, T.-H.; Dou, Q.; and Liu, Y. 2022.
\newblock AutoLaparo: A New Dataset of Integrated Multi-tasks for Image-guided Surgical Automation in Laparoscopic Hysterectomy.
\newblock In \emph{MICCAI}, 486--496.

\bibitem[{Yang et~al.(2024)Yang, Luo, Wang, and Chen}]{surgformer}
Yang, S.; Luo, L.; Wang, Q.; and Chen, H. 2024.
\newblock Surgformer: Surgical Transformer with Hierarchical Temporal Attention for Surgical Phase Recognition.
\newblock In \emph{MICCAI}.

\bibitem[{Zhang et~al.(2021)Zhang, Ghanem, Simes, Choi, Yoo, and Min}]{swnet}
Zhang, B.; Ghanem, A.; Simes, A.; Choi, H.; Yoo, A.; and Min, A. 2021.
\newblock {SWN}et: Surgical Workflow Recognition with Deep Convolutional Network.
\newblock In \emph{PMLR}, 855--869.

\bibitem[{Zisimopoulos et~al.(2018)Zisimopoulos, Flouty, Luengo, Giataganas, Nehme, Chow, and Stoyanov}]{deepphase}
Zisimopoulos, O.; Flouty, E.; Luengo, I.; Giataganas, P.; Nehme, J.; Chow, A.; and Stoyanov, D. 2018.
\newblock DeepPhase: Surgical Phase Recognition in CATARACTS Videos.
\newblock In \emph{MICCAI}, 265--272.

\end{thebibliography}

\end{document}